\pdfoutput=1
\documentclass{article}
\usepackage{spconf,amsmath,graphicx}
\usepackage{ulem}
\usepackage{xcolor}
\usepackage{CJKutf8}
\usepackage{amsfonts}
\usepackage{graphicx}
\usepackage{multirow}
\usepackage{booktabs}
\usepackage{url}
\usepackage{xcolor}

\newcommand{\MethodName}{\textsc{OneTarget}}

\title{Focus Is What You Need For Chinese Grammatical Error Correction}
\makeatletter
\def\name#1{\gdef\@name{#1\\}}
\makeatother
\name{\em{Jingheng Ye$^{1}$,
    Yinghui Li$^{1}$,
    Shirong Ma$^{1}$,} 
    Rui Xie$^{2}$, 
    Wei Wu$^{2}$,
    Hai-Tao Zheng$^{1,3*}$
\thanks{* Corresponding author. (E-mail: zheng.haitao@sz.tsinghua.edu.cn)}
}

\address{$^{1}$Shenzhen International Graduate School, Tsinghua University \\
      $^{2}$Meituan, $^{3}$Peng Cheng Laboratory}
\begin{document}
\ninept
\begin{CJK*}{UTF8}{gbsn}

\maketitle

\begin{abstract}
Chinese Grammatical Error Correction (CGEC) aims to automatically detect and correct grammatical errors contained in Chinese text. In the long term, researchers regard CGEC as a task with a certain degree of uncertainty, that is, an ungrammatical sentence may often have multiple references. However, we argue that even though this is a very reasonable hypothesis, it is too harsh for the intelligence of the mainstream models in this era. In this paper, we first discover that multiple references do not actually bring positive gains to model training. On the contrary, it is beneficial to the CGEC model if the model can pay attention to small but essential data during the training process. Furthermore, we propose a simple yet effective training strategy called \MethodName{} to improve the focus ability of the CGEC models and thus improve the CGEC performance. Extensive experiments and detailed analyses demonstrate the correctness of our discovery and the effectiveness of our proposed method.
\end{abstract}


\section{Introduction}\label{sec:intro}

Grammatical Error Correction (GEC) aims to correct all grammatical errors in a given ungrammatical text, with the constraint of keeping the original semantic as possible~\cite{wang-etal-2019-survey}. The GEC task has attracted more and more attention due to its practical value in daily life, and has been widely used in machine translation~\cite{li-etal-2021-miss}, spoken language~\cite{lu2020spoken}, and user-centric application~\cite{kaneko-etal-2022-interpretability}.
In the Chinese natural language processing community, Chinese Grammatical Error Correction (CGEC) also plays an important role~\cite{ma2022linguistic, DBLP:conf/acl/LiZLLLSWLCZ22}.

In the long-term development of CGEC, multi-reference is always an important setting that cannot be ignored because of the uncertainty and subjectivity of the CGEC task~\cite{bryant-ng-2015-far}.
Take Table~\ref{tab:intro} as an example, for the ungrammatical source sentence, there often exist two or more grammatically correct and semantically unchanged reference sentences.
We show that Lang8~\footnote{http://tcci.ccf.org.cn/conference/2018/taskdata.php}, a widely used CGEC training dataset, consists of lots of multi-reference samples, i.e., multiple different target sentences for the same ungrammatical source sentence.
In addition, the multi-reference setting is also an important aspect of the recently proposed CGEC benchmark called MuCGEC~\cite{zhang-etal-2022-mucgec}. 
\emph{But is this setting always perfect and necessary?}

\textbf{Dialectically, we argue that the multi-reference setting is reasonable for better evaluation quality, but introduces more uncertainty into the training process of the model.}
Particularly, in the process of CGEC dataset annotation, it will be difficult to decide which reference is the best among multiple equally acceptable corrections. 
Therefore, the compatibility brought by the multi-reference setting can guarantee a more realistic evaluation of the model performance~\cite{choshen-abend-2018-inherent}.
But every coin has two sides, we find that more correction uncertainty exists in multi-reference training samples, which could confuse models during training. 
We think that the main reason why the model would be confused by multiple references is that the intelligence of the mainstream models in this era is not enough to support them effectively distinguishing different references, while human language learners can easily establish complex connections between these references and promote their writing skills. 

\begin{table}[t]
\centering
\small
\renewcommand\arraystretch{1.2}
\begin{tabular}{ll}
\hline
\multicolumn{1}{c}{\multirow{2}{*}{\textbf{Source}}} & 我能胜任这此职务  \\
\multicolumn{1}{c}{} & I am competent for this the position. \\ 
\hline
\multirow{2}{*}{\textbf{Ref. 1}} & 我能胜任\textcolor{red}{这\sout{此}}职务。 \\
& I am competent for \textcolor{red}{this \sout{the}} position. \\
\multirow{2}{*}{\textbf{Ref. 2}} & 我能胜任\textcolor{red}{\sout{这}此}职务。 \\
& I am competent for \textcolor{red}{\sout{this} the} position.  \\ 
\hline
\end{tabular}
\caption{The example of CGEC. The \textcolor{red}{correction part} is marked.}
\label{tab:intro}
\end{table}

Based on the above observations and intuitions, we investigate how multi-reference samples influence the performance of CGEC models. Specifically, training models with multi-reference samples could be treated as a multi-label classification problem (for Seq2Edit-based models) or multi-target generation problem (for Seq2Seq-based models), which dilutes the prediction probability mass of models. 
Consequently, models will be confused and get into a dilemma without additional mechanisms introduced. 
In order to remedy this problem, we propose a simple yet effective training strategy called \MethodName{}, which is model-agnostic. 
\MethodName{} picks out only one reference based on different strategies for a source sentence with multiple candidate references and keeps the one-reference sample as it is. 
Though fine-tuning CGEC models with fewer references, \MethodName{} gives the models better focus ability and improves their performance, this is the meaning of our title, “focus is all you need for CGEC”.

We identify distinct advantages to training with our proposed \MethodName{}:
(1) Training models using only one-reference samples is more efficient since cleaned datasets account for 50-60\% paired samples of original datasets.
(2) Filtering unnecessary references can achieve better performance, and it holds for both the Seq2Edit-based model and the Seq2Seq-based model.

To be summarized, the contributions of our work are three-fold:
(1) We first observe and focus on the negative impact of the multi-reference setting for the training of CGEC models.
(2) We propose \MethodName{} to construct a smaller but more refined dataset, which not only speeds up training but also improves the performance of CGEC models.
(3) We conduct extensive experiments and detailed analyses on MuCGEC and achieve state-of-the-art performance.

\begin{table*}[!htb]
\centering
\scalebox{0.9}{
\begin{tabular}{lccccc}
\toprule
\textbf{Dataset} & \textbf{\# sample} & \textbf{\# err. sample (perc.)} & \textbf{\# unique source (perc.)} & \textbf{length} & \textbf{Levenshtein ratio} \\ 
\hline
Lang8 & 1,220,906 & 1,096,387 (89.8\%) & 695,275 (56.95\%) & 18.9 & 0.84 \\
HSK & 156,870 & 96,049 (61,23\%) & 156,358 (99.67\%) & 27.3 & 0.95 \\ 
MuCGEC-dev & 2,467 & 2,412 (97.77\%) & 1,137 (46.09\%) & 44.0 & 0.90 \\
\bottomrule
\end{tabular}}
\caption{Data statistics of Lang8, HSK and MuCGEC-dev, including the number of paired samples, the number (proportion) of erroneous samples, the number (proportion) of unique source sentences, the average length per source sentence and the average Levenshtein ratio between source sentences and corresponding target sentences.}
\label{tab:overall_statistics}
\end{table*}
\begin{center}
\begin{table*}[!htb]
\centering
\scalebox{0.8}{
\begin{tabular}{lccccc}
\toprule
\textbf{\# target} & \textbf{\# source (perc.)} & \textbf{length} & \textbf{mean of Levenshtein ratio} & \textbf{variance of Levenshtein ratio} & \textbf{\# edits/target}\\ 
\hline
1 & 290,493 (49.99\%) & 18.8 & 0.84 & 0.0185 & 2.1 \\
2 & 168,044 (28.92\%) & 20.1 & 0.83 & 0.0196 & 2.5 \\
3 &  73,196 (12.60\%) & 20.8 & 0.81 & 0.0202 & 2.7 \\
4 &  28,817 (4.96\%) & 21.5 & 0.81 & 0.0207 & 2.9 \\
5 &  11,507 (1.98\%) & 22.0 & 0.80 & 0.0213 & 3.0 \\
6 &   4,853 (0.84\%) & 22.3 & 0.80 & 0.0211 & 3.1 \\
7 &   2,155 (0.37\%) & 23.1 & 0.79 & 0.0216 & 3.2 \\
\textgreater 8  &  2,020 (0.35\%) & 22.7 & 0.78 & 0.0244 & 3.3 \\
\hline 
\textbf{Total} & 581,085 (100.00\%) & 19.7 & 0.82 & 0.0201 & 2.5 \\[-3pt]
\bottomrule
\end{tabular}}
\caption{Data statistics of Lang8 dataset, including the number (proportion) of unique ungrammatical source sentences (i.e., grammatical source sentences have been removed) with different number of corrected target sentences, the average length per source sentence, the average Levenshtein ratio between source sentences and corresponding target sentences, the variance of Levenshtein ratio and the average number of edits per target sentence.}
\label{tab:lang8_statistics}
\end{table*}
\end{center}

\section{Observations}\label{sec:datasets}

\textbf{We refer to “reference” as “target” in the context of training from now on.}
Recent works~\cite{zhang-etal-2022-mucgec,Zhao_Wang_2020} in the CGEC community usually train models using publicly available datasets, i.e., Lang8 and HSK
\footnote{http://hsk.blcu.edu.cn}.
Lang8 is a language learning platform, where native speakers voluntarily correct texts written by second-language learners without strict annotation regulation.
HSK (Hanyu Shuiping Kaoshi) is an official Chinese proficiency exam, of which samples come from essays written by Chinese second-language learners in the exam.
As for the recently proposed evaluation dataset, MuCGEC, a multi-reference multi-source evaluation dataset, consisting of 7,063 sentences collected from three Chinese-as-a-Second-Language learner sources
\footnote{It is worth noting that MuCGEC dataset includes 1,942 re-annotated sentences from Lang8 out of 7,063 sentences in total.}.
In order to figure out whether and how many multi-reference samples exist in public datasets, we report the statistics of Lang8, HSK and MuCGEC-dev datasets in Table~\ref{tab:overall_statistics}.
One interesting observation is that in Lang8 dataset, unique source sentences only account for 56.95\% of the number of samples while HSK consists of almost unique source sentences. Since we are interested in the impact of the number of target sentences on model training, we focus on Lang8 dataset. Further, We report in detail the statistics of Lang8 dataset in Table~\ref{tab:lang8_statistics}. 


\textbf{High correction uncertainty in multi-target samples.}
As seen in Table~\ref{tab:lang8_statistics}, as the number of target sentences increases, the number of (unique) source sentences decreases, the length of source sentences and the number of edits per target sentence grow while the Levenshtein ratio decreases with the trend.
The Levenshtein ratio is defined as follow:

\begin{equation}
L(S,T) = \frac{\textit{sum} - \textit{ldist}(S, T)}{\textit{sum}},
\end{equation}

\noindent
where $\textit{sum}$ is the sum of the length of source sentence $S$ and target sentence $T$, and $\textit{ldist}(S,T)$ is the Levenshtein distance between $S$ and $T$.

Since the Levenshtein ratio is the ratio of the Levenshtein distance and the sum of length, it could be indicative of the correction difficulty of a source sentence. The correction difficulty is a concept that can be understood more exactly as correction uncertainty.
Bryant \textit{et al.}~\cite{bryant-ng-2015-far} have shown that human annotations of GEC task are very subjective and diversified, and we further find that the correction uncertainty may be an important factor for why multi-target samples exist.
Annotators are more likely to give different corrections when facing an ungrammatical source sentence with high correction uncertainty.
As for an ungrammatical source sentence of lower correction uncertainty, on the opposite, different annotators are more likely to agree with others, and then only one unique target sentence is generated.

\textbf{Less correction uncertainty in MuCGEC dataset.}
In order to encourage more diverse and high-quality annotations, the annotation process of the MuCGEC evaluation dataset involved 21 undergraduate native students, among which 5 outstanding annotators were selected as senior annotators to review and determine final golden references~\cite{zhang-etal-2022-mucgec}.
Therefore, MuCGEC is constructed by experts and includes less noise while Lang8 is by voluntaries and includes more noise.
The Levenshtein ratio of the MuCGEC dataset (0.90 reported in Table~\ref{tab:overall_statistics}) is much higher than that of erroneous samples in the Lang8 dataset (0.82 reported in Table~\ref{tab:lang8_statistics}), showing that a proper correction should include fewer edits
\footnote{MuCGEC dataset is not annotated under minimal edit distance principle and involves more edits than original NLPCC18 dataset~\cite{zhang-etal-2022-mucgec,zhao2018overview}. Also, fluent Grammatical Error Correction is out of the scope of our work.}.
That is, the annotations of the MuCGEC dataset are of less correction uncertainty since annotations are made by experts, inflecting the goal of GEC: to correct all grammatical errors in a given ungrammatical text, with the constraint of keeping the original semantic of the sentence as possible.

\section{Methodology}\label{sec:method}

\subsection{Multi-target Training}\label{subsec:multi-target}

In this section, we analyze what if we train models using multi-target samples.
Without the loss of generalization, we take Seq2Edit-based models as examples
\footnote{The discussion of Seq2Seq-based models is similar but more tedious due to auto-regression.}.
Seq2Edit-based models treat GEC as a sequence tagging problem and decode tag sequences in parallel. Owing to the existence of multi-target samples, GEC is essentially a multi-label classification problem.
As shown in Table~\ref{tab:intro}, when coming to tokens ``这'' and ``此'', cross-entropy loss responding to Ref. 1 encourages models to keep ``这'' and delete ``此'' while Ref. 2 takes the opposite.
If considering simultaneously both equally reasonable corrections, the tokens ``这'' and ``此'' correspond to multiple valid candidate labels, i.e., \textit{delete} or \textit{keep}, thus diluting models' probability mass, as a result of which models get confused and the performance drops.
Therefore, in the setting of multi-target training, different corrections may be contradictory but the intelligence of the mainstream models in this era is not enough to support them effectively understanding different corrections.




\subsection{One-target Training}\label{subsec:one-target}
To remedy the above problem caused by multi-target training, we propose a simple yet effective data cleaning method called \MethodName{}. We pick out a single target sentence based on three strategies for a source sentence with multiple target sentences.
Recent works~\cite{lee-etal-2022-deduplicating,mishra-sachdeva-2020-need} show that removing duplicates during pre-training can significantly increase training efficiency with equal or even better performance. Our work, however, removes redundant target sentences for the same source sentence, which is a customized deduplication step for the GEC task.
Specifically, a single target sentence is selected


\noindent(1) \textbf{Based on Levenshtein ratio}. For a source sentence with multiple feasible target sentences, we compute the Levenshtein ratio respectively between each target sentence and the same source sentence. Samples with the most Levenshtein ratio are denoted as \textbf{lev\_sim} and samples with the least Levenshtein ratio are denoted as \textbf{lev\_dis}. Please refer to the definition of the Levenshtein ratio in Section~\ref{sec:datasets}.




\noindent(2) \textbf{Based on Jaccard similarity}. Jaccard similarity is used for gauging the similarity and diversity of token sets. Similarly, samples with the most Jaccard score are denoted as \textbf{jac\_sim} and samples with the least Jaccard score are denoted as \textbf{jac\_dis}. Jaccard similarity is defined as follow:

\begin{equation}
J(S,T) = \frac{|S\cap T|}{|S\cup T|},
\end{equation}

\noindent
where $|S\cap T|$ is the number of tokens that appear in both $S$ and $T$ and $|S\cup T|$ is the number of tokens that appear in either $S$ or $T$.

\noindent(3) \textbf{Based on the number of edits}. The above two strategies are literally computed and introduce no linguistic information. To explore whether and how target sentences with different numbers of edits influence the performance of models, we first convert paired data to M$^2$ format using ERRANT~\cite{bryant-etal-2017-automatic}, a grammatical error annotation toolkit designed to automatically extract edits 
\footnote{We use ERRANT-zh~\cite{zhang-etal-2022-mucgec} for CGEC.}, which takes into consideration linguistic information~\cite{felice-etal-2016-automatic}.
Samples with the least number of edits are denoted as \textbf{edi\_least} and samples with the most number of edits are denoted as \textbf{edi\_most}.

\section{Experiment}\label{sec:experiments}

In order to demonstrate explicitly the effectiveness of our proposed data cleaning method, we compare GEC models trained with different \MethodName{} strategies on MuCGEC-dev and then evaluate our best models on MuCGEC-test.

\subsection{Experimental Setup}\label{subsec:setup}

\noindent\textbf{Datasets and Metrics.}
We first limit our training data to Lang8 dataset if not mentioned otherwise, since HSK dataset does not consist of many multi-target samples.
Following Zhang \textit{et al.}~\cite{zhang-etal-2022-mucgec}, we filter duplicate sentences in MuCGEC dataset for real performance and also discard correct sentences. We adopt ERRANT-zh proposed by Zhang \textit{et al.}~\cite{zhang-etal-2022-mucgec} as our evaluation metric and evaluate models on MuCGEC dataset.

\noindent\textbf{Models.}
To show that our proposed method is unrelated to the model structure, we train respectively GECToR~\cite{omelianchuk-etal-2020-gector} as Seq2Edit-based models, which achieves the SOTA performance on EGEC datasets, and BART~\cite{lewis-etal-2020-bart} as Seq2Seq-based models. Following Zhang \textit{et al.}, we initialize our Seq2Edit-based models with StructBERT~\cite{wang2019structbert} and Seq2Seq-based models with Chinese-BART~\cite{shao-etal-chinese-bart}. Since some common characters are not contained in the vocabulary of Chinese-BART, we extend its vocabulary. We also train GEC models using \textbf{random} one-target dataset and \textbf{full} original dataset for comparison with our data cleaning method introduced in Section~\ref{subsec:one-target}.

\noindent\textbf{Hyper-parameters.}
We use AdamW optimizer with learning rate 1e-5, batch size 32, and gradient accumulation step 1.
We set the number of training epochs to 10 for all experiments and evaluate models every epoch.
The other hyper-parameters are the same with publicly available code~\cite{zhang-etal-2022-mucgec}.
All experiments are conducted for 4 runs and the averaged metrics are reported.

\subsection{Results on MuCGEC-dev}\label{subsec:dev}

\begin{table}[!tbp]
\centering
\scalebox{0.8}{
\begin{tabular}{l|l|ccc|ccc}
\toprule
\multicolumn{2}{c|}{\multirow{2}{*}{\textbf{Strategy}}} & \multicolumn{3}{c|}{Seq2Seq} & \multicolumn{3}{c}{Seq2Edit} \\
\multicolumn{2}{c|}{} & \textbf{P} & \textbf{R} & \textbf{F}$_{0.5}$ & \textbf{P} & \textbf{R} & \textbf{F}$_{0.5}$ \\ 
\midrule
\multirow{8}{*}{\MethodName{}} & \textbf{lev\_sim} & 46.14 & 27.30 & \textbf{40.55} & \textbf{47.20} & 16.60 & \textbf{34.48} \\
& \textbf{lev\_dis} & 43.72 & 27.96 & 39.29 & 38.38 & 18.34 & 31.50 \\ 
\cmidrule{2-8}
& \textbf{jac\_sim} & 46.25 & 26.71 & \textbf{40.35} & 45.75 & 15.63 & 33.02 \\
& \textbf{jac\_dis} & 43.68 & 26.33 & 38.59 & 38.77 & 17.73 & 31.34 \\ 
\cmidrule{2-8}
& \textbf{edi\_least} & \textbf{46.33} & 24.69 & 39.42 & 43.90 & 16.30 & 32.80 \\
& \textbf{edi\_most}  & 44.04 & 27.18 & 39.18 & 37.25 & \textbf{21.21} & 32.25 \\ 
\cmidrule{2-8}
& \textbf{random} & 45.45 & 25.44 & 39.27 & 41.42 & 17.27 & 32.37 \\
\midrule
\multicolumn{2}{c|}{\textbf{full}} & 44.08 & \textbf{28.16} & 39.60 & 41.36 & 19.18 & 33.59 \\ 
\bottomrule
\end{tabular}}
\caption{Performance on MuCGEC-dev of Seq2Seq-based and Seq2Edit-based models with different strategies.}
\label{tab:exp1}\end{table}

Table~\ref{tab:exp1} reports the performance on MuCGEC-dev of models trained with different \MethodName{} strategies. Please note that the number of training samples is 581,085 for \MethodName{} training and 1,096,387 for full multi-target training.

\textbf{Comparison between \MethodName{} training and multi-target training.} Surprisingly, although the number of paired samples is approximately half of the full dataset, both Seq2Seq-based and Seq2Edit-based models trained with \MethodName{} achieve equal or even better performance. Please note that \MethodName{} does not introduce additional improvement on the model structure. It shows that training models with multi-target samples depart from Pareto optimality of both efficiency and performance.

\textbf{Comparison between different \MethodName{} strategies.}
Although different \MethodName{} strategies lead to the same number of samples, both Seq2Seq-based and Seq2Edit-based models trained with different \MethodName{} strategies achieve distinct performance and reveal similar trends.
Firstly, the strategy \textbf{lev\_sim} performs best among our proposed three strategies, followed by \textbf{jac\_sim}, \textbf{edit\_least}, \textbf{full} and \textbf{random} in order.
In addition, \textbf{*\_sim} and \textbf{edit\_least} strategies perform better than \textbf{random}, \textbf{*\_dis} and \textbf{edit\_most}. This trend suggests that GEC prefers corrections without many edits, providing insight into GEC dataset construction. Surprisingly, \textbf{edit\_least} performs lightly better than \textbf{random} but falls behind \textbf{lev\_sim} and \textbf{jac\_sim}. A possible reason is that the number of edits could not reflect the data quality.

Another interesting observation is that \MethodName{} can significantly increase the Precision without a great decrease in the Recall. 
Specifically, \textbf{random} and \textbf{full} settings result in models' modest Precision and Recall, while higher precision and lower Recall for \textbf{*\_sim} or \textbf{edit\_least}, and lower Precision and higher Recall conversely for \textbf{*\_dis} or \textbf{edit\_most}. This phenomenon can be explained through the trade-off between the quality and the number of edits. For example, \textbf{lev\_sim} prefers paired samples whose target sentences are more similar to source sentences in terms of Levenshtein ratio. Even if target sentences selected in this way contain fewer edits, it is assumed that source sentences are corrected reasonably, therefore in this case it filters out those paired samples with many unnecessary edits. That is, the quality and the number of edits play an important role, such that training models with less yet high-quality edits improve the Precision with a decrease in the Recall, and vice versa. We recommend choosing samples whose target sentences are similar to source sentences since Precision matters more than Recall for the CGEC task.

\begin{table}[!tbp]
\centering
\scalebox{0.8}{
\begin{tabular}{l|ccc|ccc}
\toprule
\multicolumn{1}{c|}{\multirow{2}{*}{\textbf{}}} & \multicolumn{3}{c|}{Seq2Seq} & \multicolumn{3}{c}{Seq2Edit} \\
\multicolumn{1}{c|}{} & \textbf{P} & \textbf{R} & \textbf{F}$_{0.5}$ & \textbf{P} & \textbf{R} & \textbf{F}$_{0.5}$ \\ 
\midrule
\textbf{Zhang et al.} \cite{zhang-etal-2022-mucgec} & 43.81 & 28.56 & 39.58 & 44.11 & 27.18 & 39.22 \\
\textbf{Zhang et al.} $\clubsuit$ & 46.49 & 28.64 & \textbf{41.34} & 46.62 & 25.60 & 40.04 \\
\midrule
\textbf{lev\_sim} & 45.95 & 28.87 & 41.09 & 47.34 & 25.97 & \textbf{40.65} \\
\textbf{random} & 47.14 & 27.28 & 41.15 & 44.66 & 27.55 & 39.73 \\ 
\bottomrule
\end{tabular}}
\caption{Comparison of our single models with related work on MuCGEC-test. We only report part of the results due to the limitation of space. Models in the first block are trained using the full dataset (with 1,557,438 paired samples) and models in the second block are trained using one-target dataset (with 1,056,600 paired samples). The symbol of “$\clubsuit$” means the results of our re-implemented baseline method.}
\label{tab:mucgec_test}\end{table}

\subsection{Results on MuCGEC-test}\label{subsec:test}

In this section, we evaluate our best trained models on MuCGEC-test, selecting \textbf{lev\_sim} strategy and \textbf{random} for comparison.
Following Zhang \textit{et al.}~\cite{zhang-etal-2022-mucgec}, we duplicate the HSK data five times and merge them with Lang8 data. The results reported in Table~\ref{tab:mucgec_test} show that our models achieve comparable or even better F$_{0.5}$ scores with fewer training samples. Please note that the HSK dataset consists of almost one-target samples and samples in HSK are duplicated five times, so the improvement is limited compared to Table~\ref{tab:exp1}.

Additionally, it is worth mentioning that our proposed method has won the first prize in the CCL2022-CLTC competition~\footnote{https://tianchi.aliyun.com/dataset/dataDetail?dataId=131328} based on MuCGEC-test, which demonstrates the effectiveness and competitiveness of \MethodName{}.

\subsection{Ablation Study: Influence of the Target Number}\label{subsec:exp2}

In order to further study the impact of the target number on the model performance, we train models using three artificial datasets with a different number of target sentences but the same number of source sentences.
Specifically, we pick out source sentences with three or more target sentences randomly in the Lang8 dataset, and a total of 127,986 unique source sentences are obtained. For each source sentence, we construct 1$\sim$3-target datasets, such as 2-target dataset consisting of 127,986$\times$2=255,972 paired samples. Note that each of these datasets includes the same number of unique source sentences regardless of n.
Due to the limitation of space, we fine-tune three Seq2Edit-based models with 1$\sim$3-target datasets respectively and report the results on MuCGEC-dev in Table~\ref{tab:exp2}.

\begin{table}[!htpb]
\centering
\scalebox{0.8}{
\begin{tabular}{llccc}
\toprule
\textbf{Dataset} & \textbf{\#sample} & \textbf{P} & \textbf{R} & \textbf{F$_{0.5}$} \\ 
\hline
1-target & 127,986 & \textbf{42.03} & 15.67 & \textbf{31.45} \\ 
2-target & 255,972 & 37.50 & \textbf{18.51} & 31.11 \\
3-target & 383,958 & 39.22 & 16.74 & 30.92 \\ 
\bottomrule
\end{tabular}}
\caption{Performance on MuCGEC-dev of GECToR models trained with 1$\sim$3-target datasets respectively. For the sake of simplicity, we do not introduce pre-training on synthetic parallel dataset and multi-stage fine-tuning as Omelianchunk \textit{et al.} \cite{omelianchuk-etal-2020-gector}.}
\label{tab:exp2}
\end{table}

It is clear that 1-target training is the best among 1$\sim$3-target training settings, which is counter-intuitive since more training samples should have achieved better performance. The main reason is that 1-target training results in higher Precision without a great decrease in the Recall, which caters to the need for grammar error correction that pays more attention to Precision.

These phenomena are interesting and can be understood after considering the features of 1$\sim$3-target datasets. On the one hand, though with a different number of samples, these datasets consist of the same number of unique ungrammatical sentences, as a result of which all these models observe the same grammatical errors while multi-target datasets provide multiple equally reasonable corrections for the same ungrammatical sentence.
Not like humans, however, models cannot fully understand these multiple corrections, and these different corrections confuse models instead.

The fact that more samples decrease the performance of models instead inspires us to think about the relationship between grammatical error correction and training datasets. More training samples do not necessarily improve the performance of models, though more unique source sentences can indeed improve the performance of models. For example, Seq2Edit-based models trained with \MethodName{}-random strategy (581,085 samples) achieve 32.37 F$_{0.5}$ score shown in Table \ref{tab:exp1} while they achieve 31.45 F$_{0.5}$ score using 127,986 \MethodName{}-random training samples shown in Table \ref{tab:exp2}.


\section{Conclusion}\label{sec:conclusion}

In this paper, we observe that more correction uncertainty exists in multi-reference training samples. 
Though multi-reference is necessary for evaluation, it is a heavy burden for model training instead in this era.
Therefore, we come up with a simple yet effective data training strategy called \MethodName{}, which reduces the number of training samples significantly with equal or even better performance.
We further demonstrate that multi-reference training samples improve the Recall of models at the cost of a great decrease in Precision, which is not recommendable for CGEC.
In the future, we will explore establishing multi-reference relationships for models, enabling models to understand exactly the essence of CGEC.

\label{sec:refs}


\bibliographystyle{IEEEbib}
\bibliography{strings,refs}
\end{CJK*}
\end{document}